\pdfoutput=1

\documentclass[11pt]{article}

\usepackage{acl}

\usepackage{times}
\usepackage{latexsym}
\usepackage{xcolor}

\usepackage[T1]{fontenc}

\usepackage[utf8]{inputenc}

\usepackage{microtype}

\usepackage{inconsolata}

\usepackage{float}
\usepackage{amsmath}
\usepackage{amsfonts}
\usepackage{xcolor}
\usepackage{booktabs}
\usepackage{multirow}
\usepackage{array,tabularx}
\usepackage{booktabs}
\usepackage{graphicx}
\usepackage{subfig}

\usepackage{fancyvrb}
\DefineVerbatimEnvironment{verbatim}{Verbatim}{xleftmargin=.18in}

\title{Large Language Models Need Consultants for Reasoning: Becoming an Expert in a Complex Human System Through Behavior Simulation}

\author{Chuwen Wang, Shirong Zeng \and Cheng Wang\\
  Department of Computer Science and Technology, Tongji University, Shanghai, China \\
  \texttt{\{hakys, 2051857, chengwang\}@tongji.edu.cn} \\}

\begin{document}
\maketitle
\begin{abstract}
Large language models (LLMs), in conjunction with various reasoning reinforcement methodologies, have demonstrated remarkable capabilities comparable to humans in fields such as mathematics, law, coding, common sense, and world knowledge. In this paper, we delve into the reasoning abilities of LLMs within complex human systems. We propose a novel reasoning framework, termed ``Mosaic Expert Observation Wall'' (MEOW) exploiting generative-agents-based simulation technique. In the MEOW framework, simulated data are utilized to train an expert model concentrating ``experience'' about a specific task in each independent time of simulation. It is the accumulated ``experience'' through the simulation that makes for an expert on a task in a complex human system. We conduct the experiments within a communication game that mirrors real-world security scenarios. The results indicate that our proposed methodology can cooperate with existing methodologies to enhance the reasoning abilities of LLMs in complex human systems.
\end{abstract}

\section{Introduction}

Large language models (LLMs) are evolving rapidly. Through extensive training on vast datasets, they exhibit remarkable open-domain reasoning capabilities. Llama-2 \cite{touvron2023llama} shows superior reasoning performance across various evaluation benchmarks, including mathematics \cite{cobbe2021training,hendrycks2measuring}, coding \cite{chen2021evaluating}, common sense \cite{bisk2020piqa, sap-etal-2019-social}, world knowledge \cite{kwiatkowski-etal-2019-natural}, and reading comprehension \cite{rajpurkar2018know,choi-etal-2018-quac}. Techniques such as fine-tuning (FT) \cite{hu2021lora, roziere2023code}, chain-of-thought (CoT) \cite{kojima2022large}, in-context learning (ICL) \cite{wei2022chain}, and retrieval-augmented generation (RAG) \cite{lewis2020retrieval} further enhance the performance of LLMs in specific tasks or domains. However, these methodologies may not deal well with complex human strategic decision-making and interaction scenarios. For instance, in the context of financial security, due to privacy issues, there is a lack of data and explicit knowledge to implement FT and RAG effectively. Moreover, ICL combined with CoT often leads LLMs to stick to examples or historical information when solving problems. In fact, this approach is not always effective, as financial attackers may not repeat previously detected or blocked methods to attack the financial system, and even some valid behaviors by the normal can lead to systemic risk \cite{eisenberg2001systemic}.

Simulation techniques serve as indispensable tools for tackling challenges within complex human systems. An inherent feature of simulation is its capacity to delve into interactions among individuals, offering an optimal solution to problems related to complex systems. Studies across diverse fields \cite{5644908, zhou2020multi, hui2022knowledge} have validated their effectiveness. With the advent of generative agents technique \cite{generative} which realized individual-level human behavior simulation in a sandbox, we believe that simulation techniques based on it can indeed extend beyond complex natural systems to complex human systems.

\begin{figure}
	\centering
	\includegraphics[width=0.45\textwidth]{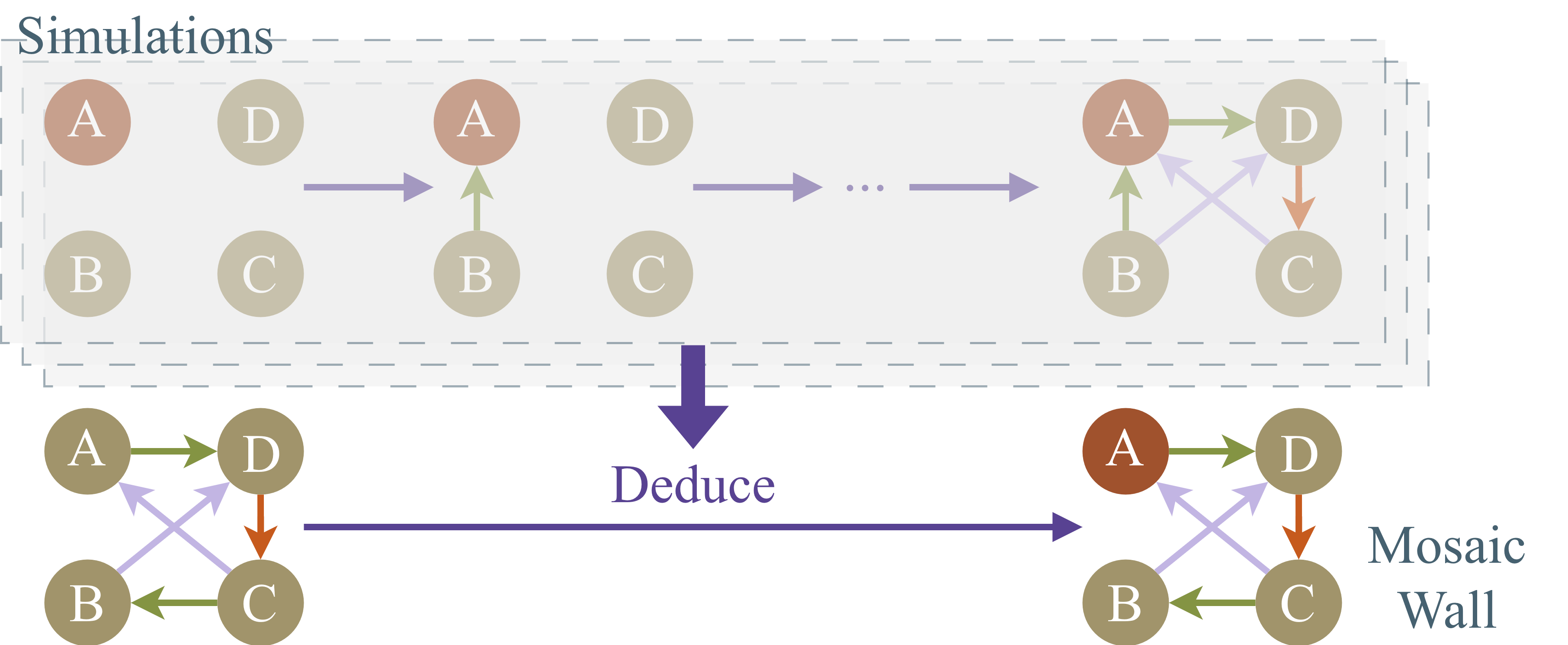}
	\caption{Mosaic investigation wall is what detectives use to visualize clues and interactions within cases. Through simulating current individuals' behaviors and interactions on a Mosaic wall, a detective can deduce who may be the crime in corresponding real case.}
	\label{fig:MW}
	\vspace{-10pt}
\end{figure}

In this paper, after utilizing the generative-agents-based simulation technique to simulate a complex human system, we design a novel framework, ``Mosaic Expert Observation Wall'' (MEOW), which imitates a detective simulating and analyzing a case through Mosaic investigation wall as illustrated in Figure \ref{fig:MW}. In MEOW, real game data are processed by an expert model trained on simulated data and converted into natural language prompts as expert observation to assist LLM reasoning. From the perspective of assisted LLM agent, the expert model serves as its consultant. This novel methodology addresses aforementioned problems of existing LLM reasoning reinforcement methodologies, which offers the potential to improve LLM reasoning independently or in conjunction with them in complex human systems. As a validation, we conduct the experiments in a Werewolf-like communication game where players infer others' identities based on common sense and game strategies. This scenario is a simplified representation of real-world security scenarios involving conscious attackers and involuntary negative behaviors. Subsequently, another LLM-based agent acts as a detective to infer each player's identity. 

MEOW addresses challenges in analyzing problems in complex human systems and is capable of collaborating with existing LLM reasoning reinforcement approaches. Our experimental results demonstrate the effectiveness of MEOW. Additionally, through these experiments, we stumble across and summarize some serious challenges of applying MEOW to more complex human systems.

\section{Related Work}

\subsection{LLM Reasoning}

When applying a general LLM to a specific scenario, there is often a demand for focusing on knowledge relevant to that scenario. To meet this demand, methodologies such as FT, CoT, ICL, and RAG have been proposed and are widely used. Among them, FT is the most powerful one. Efficient fine-tuning methods based on Low-Rank Adaptation of LLMs (LoRA) \cite{hu2021lora} have enabled the realization of specialized LLMs for code \cite{roziere2023code}, biomedical applications \cite{tinn2023fine}, and more. CoT \cite{kojima2022large}, originating from a special prompt, ``Let’s think step by step'', is a data-free approach to improve LLM reasoning on complex problems. CoT methods, and further tree-of-thought \cite{yao2023tree} feature using special prompts and context organization modes to direct the LLM in generating a series of steps to solve a problem. Since LLMs require context to filter candidate tokens, more sufficient and meaningful contexts enhance their ability to reason about complex problems. ICL, which makes LLMs learn from analogy, was first applied to solve mathematical problems \cite{wei2022chain}. Through prompts of problem-solving examples, the LLM imitates the correct steps, making it a kind of manual CoT based on example data \cite{dong2022survey}. RAG, leveraging external professional knowledge, can effectively mitigate the hallucination of LLMs \cite{lewis2020retrieval}. By retrieving external knowledge, LLM applications like Copilot \footnote{\href{https://copilot.microsoft.com/}{Copilot: https://copilot.microsoft.com/}} and GPT-4 \cite{achiam2023gpt} can generate and reason more accurately. However, FT and RAG require extremely huge amounts of high-quality data that are often unavailable in complex human systems. While CoT and ICL require less data, the complexity of human systems makes it challenging to solve problems using analogical and simple ``step by step'' patterns. Therefore, none of these methodologies is good enough to analyze complex human systems at present.

\subsection{Generative-agents-based Simulation}

Generative-agents-based simulation is a form of agent-based simulation whose agents are text-based LLMs instead of typical deep learning (DL) models trained in closed domains. Proposed by \citet{generative}, this simulation leverages the strengths of generative agents, encompassing common sense reasoning, contextual retrieval, and humanity. With these advantages, logical thoughts and human-like social behaviors have been observed during the simulation in a sandbox. Furthermore, \citet{gao2023s} developed a social network simulation framework building upon the work of generative agents. Their experiments demonstrated that both individual-level and population-level simulations within the generative agents social network had remarkable alignment with data of information, emotion, and attitude propagation in real world.

\begin{figure*}[t]
	\centering
	\includegraphics[width=0.8238\textwidth]{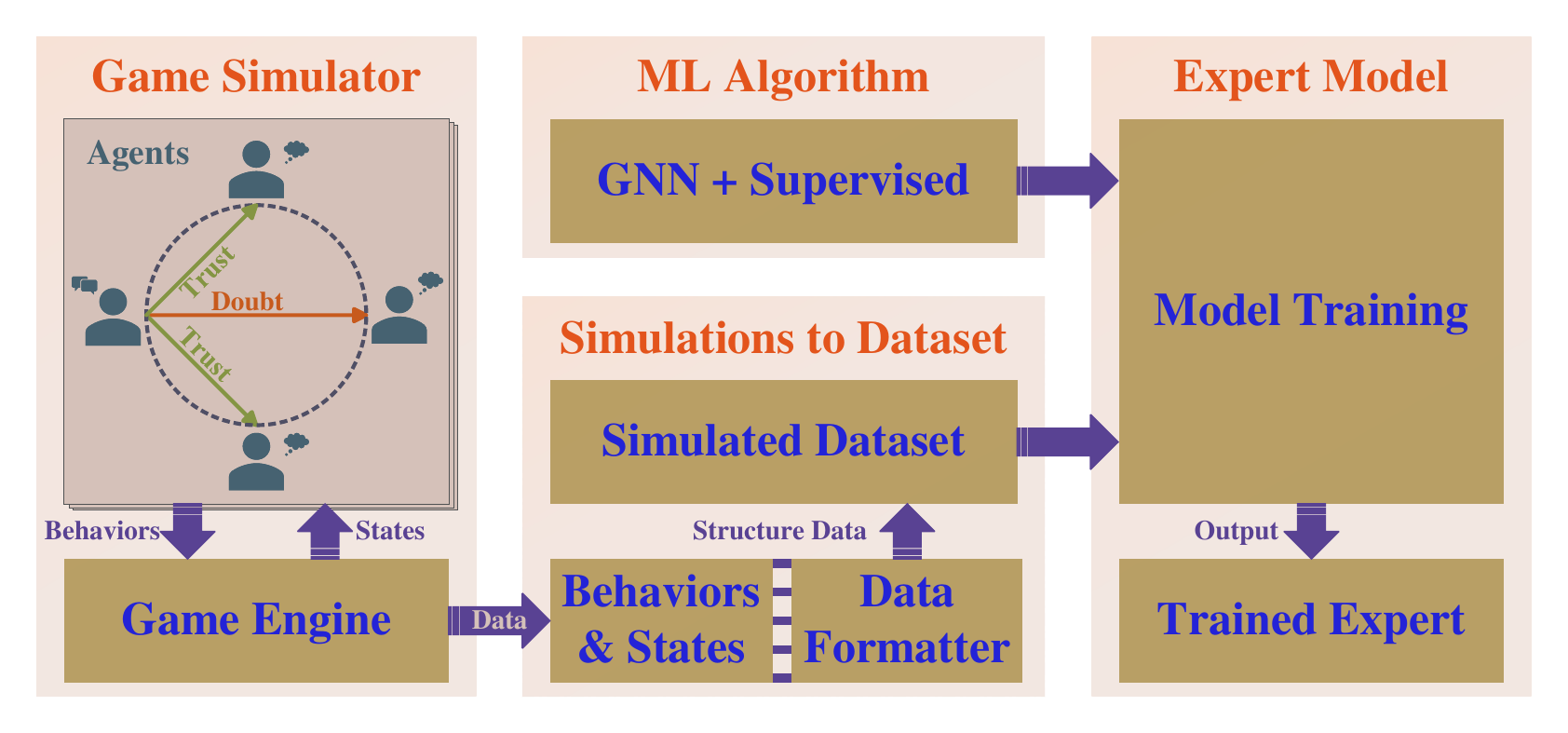}
	\caption{The illustration of implementing generative-agents-based simulation and training an expert model based on simulated data in the four-player version of ``Find The Spy''.}
	\label{fig:sys}
\end{figure*}

\section{Behavior Simulation in Communication Game}

\subsection{Game Rules}

The communication game under discussion in this study is called ``Find The Spy''. Our experiments are based on a four-player version of this game. In this scenario, players are divided into two groups: ``folk'' and ``spies''. Three players belong to the ``folk'' group, while one player is designated as a ``spy''. At the beginning of the game, ``folk'' players receive the same word, while the ``spy'' receives a different word. These two words, while distinct, share some commonalities (e.g. two words, ``apple'' and ``pineapple'', are both fruits). Each player only knows his own word and remains ignorant of both his identity (``ordinary people'' belongs to ``folk'' or ``spy'' belongs to ``spies'') and the identities of the other players.

The objective of the game is to eliminate the opposing group of players through communication and voting. For ``spies'', the goal is to conceal their identity to avoid being voted out. In the four-player scenario, when only two players remain and one of them is the ``spy'', the referee announces the ``spies'' as the winners. Conversely, the ``folk'' players aim to identify the ``spies'' based on the communication and vote them out through the voting process. If all remaining players are ``folk'', the referee announces the ``folk'' as the winners.

The detailed game processes are documented in Appendix \ref{sec:GP}.

\subsection{The Scope of Behavior Simulation}

Due to the features of communication games, players are required to possess basic common sense, reasoning abilities, and expressive skills. Tiny differences in the comprehension and analyses of utterances may lead a game to another result. It is these uncertain factors that render the game engaging and unpredictable, thereby constituting what we refer to as a complex human system. As the essence of complex systems, neither closed-form expressions nor accurate distributions about future states are feasible due to the chaotic interactive relationships among individuals in such systems \cite{lorenz1963deterministic, may1976simple}. Consequently, there is no fixed distribution of individual behaviors and states in a complex human system, implying that an increased volume of data will facilitate a more comprehensive analysis of this system, if the data differ from existing historical data. To acquire such data in a complex human system, direct observation of the real world is the optimal method, but often costly, inefficient, time-consuming, and even unavailable. Under such circumstances, resorting to simulated data emerges as an alternative. DL generative models have proven their effectiveness in supplementing image datasets for better training \cite{shrivastava2017learning}. However, relying exclusively on the outputs of DL models may not be effective in a complex human system. On the one hand, images within a certain dataset follow a relatively stable distribution compared to behaviors and states in a complex human system, meaning that supplementing the data with data of the same distribution contributes to a specific image dataset but not to a complex human system. On the other hand, the complexity of behaviors and states, coupled with limited real-world historical data, makes it challenging to train a DL model. As a result, behavior simulation currently stands as the only method to extend or even create data to assist analyzing a complex human system. Fortunately, with LLMs, it is feasible to create a generative agent capable of playing this game with sufficient knowledge as behavior simulation in the game system.

\subsection{Instance in ``Find The Spy''}

Specifically for the ``Find The Spy'' game, we design an LLM-driven judge agent tasked with identifying the ``spy'' through players' utterances and votes, relying solely on knowledge acquired from pre-training. Simultaneously, given that no historical data are available, we utilize behavior simulation to create simulated ``Find The Spy'' game datasets, which are used to train an expert model that predicts the ``spy'' player. We name this framework the ``Mosaic Expert Observation Wall'' (MEOW). In MEOW, the simulated data, rich in agents' interaction behaviors and states, constructs multiple graph data as mosaic investigation walls. Expert machine learning (ML) models, trained on these data, provide the judge agent with expert observations on the real system to refine the analyses. We anticipate that the judge agent's performance in identifying the ``spy'' will improve after consulting the expert models. If such improvements are observed, it indicates that LLM agents can reason more effectively in complex human systems through simulation, a data-efficient method besides CoT and FT.

\section{Architecture and Methodology}

To implement MEOW in the communication game, as illustrated in Figure \ref{fig:sys}, a game simulator is required to perform a specified number of simulations. Subsequently, the simulated data of behaviors and states are converted from text into graph-structured data. Finally, expert models are trained on the data and labels, using the chosen model and algorithm. In the following subsections, we will elaborate on the construction of the game simulator and the detailed implementation of MEOW.

\subsection{Game Simulator}

Our game simulator primarily comprises two components: generative agents and a game engine. The generative agents, driven by multi-turn dialogues, generate their analyses, utterances, and behaviors, including tendencies and votes, in text form. These behaviors are converted into dictionary-format data, which are then inputted into the game engine to represent interactions among agents. Subsequently, the game engine processes these interactions and updates the states of the system and agents in accordance with the game rules. Finally, upon the conclusion of the game, all behaviors and states data are collected and reformatted as the output of the simulator.

\subsubsection{LLM Agent}

A generative agent takes action based on its memory, planning, retrieval, and reflection \cite{generative}. In the ``Find The Spy'' game, our player agents are designed in a similar paradigm. We utilize CoT prompt templates to guide the inference of player agents during different phases of the game. We divide the game into three processes and six phases, which are described in detail in Appendix \ref{sec:SP}. The complete prompts and source code are available on github \footnote{\href{https://github.com/HaKyS-A/MEOW}{https://github.com/HaKyS-A/MEOW}}.

\subsubsection{Game Engine}

The game engine primarily serves two functions. The first function involves broadcasting open information. Given that LLM agents heavily rely on context, particularly in multi-turn dialogue role-play scenarios, we incorporate confirmation responses to ensure that the agents receive the broadcast successfully. For instance, 
\begin{verbatim}[fontsize=\small, commandchars=\\\{\}]
\textbf{User}: (system)(Player Bob is eliminated.)
\textbf{Assistant}: Okay, I see.
\end{verbatim}
This context is manually added instead of being generated by the LLM.

The second function of the game engine is to update the states of the system and agents, such as counting votes and eliminating. Concurrently, it records the utterances and behaviors in files as the simulator output. These files are subsequently transformed during the implementation of MEOW.

\subsection{Implementation of MEOW}

\begin{figure}
	\centering
	\includegraphics[width=0.45\textwidth]{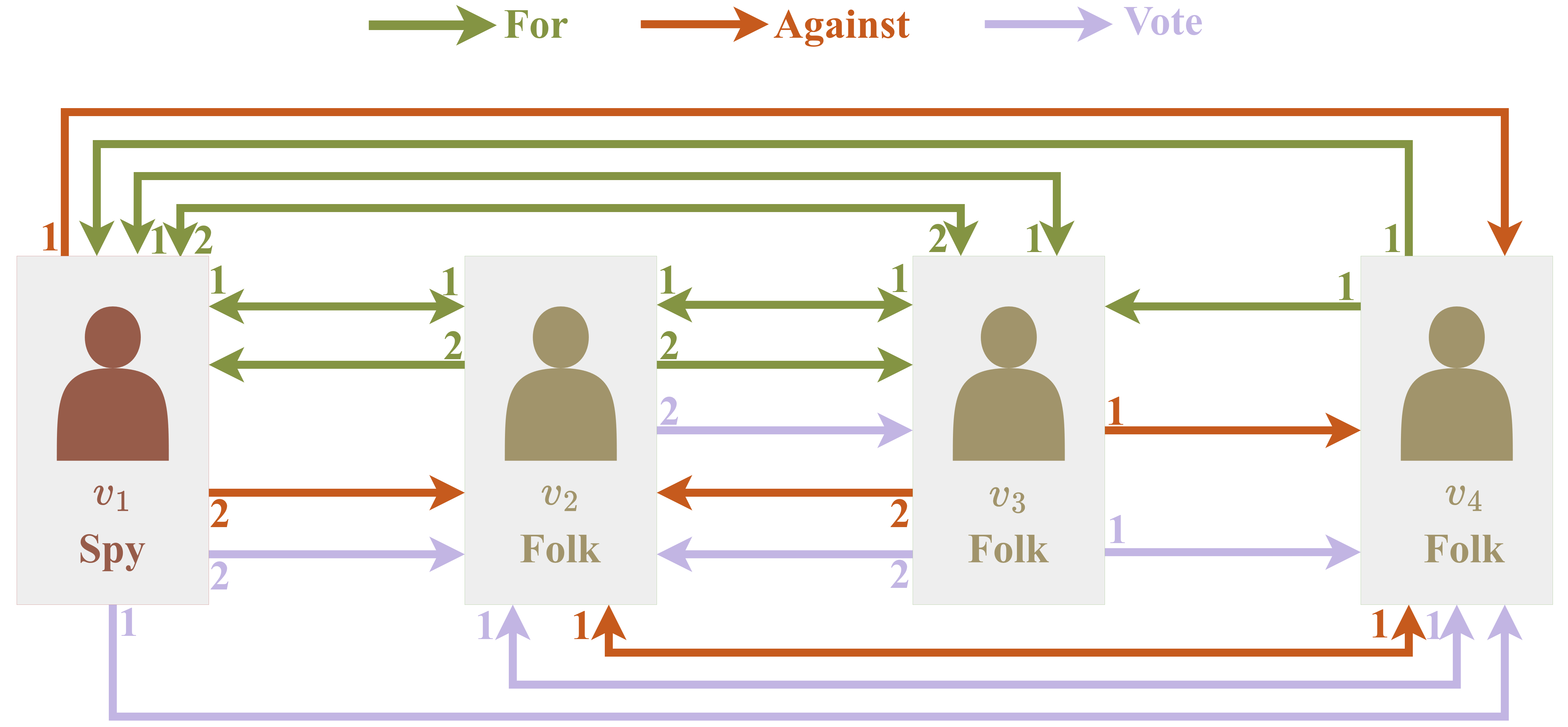}
	\caption{Example of heterogeneous graph. The numerical labels $i$ at the tail of each arrow represent the $i$th round of the game.}
	\label{fig:HeteroData_examples}
\end{figure}

Once acquiring the simulated data, the initial steps involve transforming this data into graph-structured representations and generating datasets. Utilizing these datasets, we train two expert models. During the inference phase of the actual game, the judge agent identifies the ``spy'', initially depending on raw record files, and then the inference is refined with expert observations from the trained expert models.

\subsubsection{Simulated Data to Heterogeneous Graph} \label{sec:DC}

The simulated data are transformed into heterogeneous graphs, as depicted in Figure \ref{fig:HeteroData_examples}. Each game corresponds to a directed heterogeneous graph, denoted as $G=(V,E)$, where $V$ represents the set of nodes and $E$ denotes the set of directed edges. The set $V$ comprises a single type of node, i.e., 
\begin{equation*}
	V=\lbrace v_i|(\mathbf{x}_i,\mathbf{y}_i) \rbrace_{i=1}^{n},\mathbf{x}_i \in \mathbb{R}^{1 \times 4},\mathbf{y}_i \in \lbrace 0,1 \rbrace,
\end{equation*}
where each node $v_i$ corresponds to a player and $n$ signifies the total number of players (in our experiments, $n=4$). Node $v_i$ is characterized by a 4-dimensional one-hot tensor feature $\mathbf{x}_i$ and a label $\mathbf{y}_i$ indicating its role (0 for ``folk'' and 1 for ``spy''). One game consists of either one or two rounds. The set of directed edges are denoted as:
\begin{equation*}
\begin{aligned}
	&E_1=E_{for}^1 \cup E_{against}^1 \cup E_{vote}^1, \\
	&E_2=E_1\cup E_{for}^2 \cup E_{against}^2 \cup E_{vote}^2, \\
	&E_1, E_2\subseteq V\times V,
\end{aligned}
\end{equation*}
where $E_{for}^i$, $E_{against}^i, E_{vote}^i$ represent the trusting, doubting, and voting interactions of the top $i$ rounds respectively, if the round exists.

Denoting $\mathbf{G}$ as the set of all the games and $\mathbf{G}_\lambda$ as the set of games where ``spy'' is not eliminated in the first round. Then, we create two datasets $D_1, D_2$ respectively containing the first round of each game in $\mathbf{G}$, and two rounds of partial games in $\mathbf{G}_\lambda$. 
\begin{equation*}
\begin{aligned}
	&D_{1}=\lbrace G_k | G_k=(V^k,E_{1}^k), G_k \in \mathbf{G}\rbrace, \\
	&D_{2}=\lbrace G_l | G_l=(V^l,E_{2}^l), G_l \in \mathbf{G}_\lambda\rbrace. \\ \\
\end{aligned}
\end{equation*}

\subsubsection{Expert Model Design}\label{sec:sec_expert_model}

\begin{figure}
	\centering
	\includegraphics[width=0.45\textwidth]{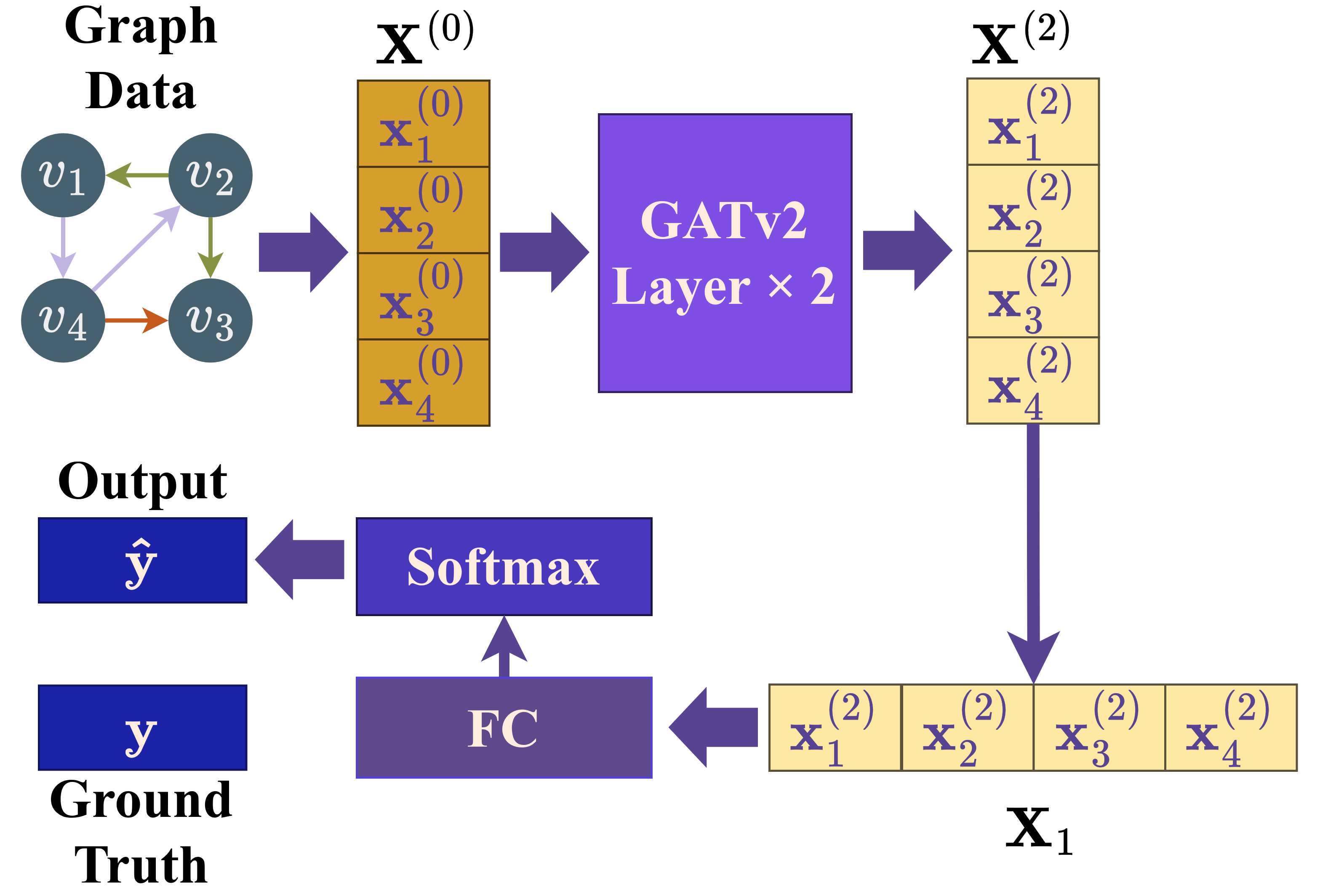}
	\caption{The architecture of expert models.}
	\label{fig:model}
	\vspace{-10pt}
\end{figure}

We formulate the problem as a four-class classification task, where one out of the four players is the ``spy''. To address this problem, we introduce two expert models respectively trained on $D_{1}$ and $D_{2}$. These models share a similar architectural design, as illustrated in Figure \ref{fig:model}. We take the first expert model as an example to illustrate its design.

The model employs two layers of the Graph Attention Network v2 (GATv2) \cite{brody2021attentive}. In the one-round graph that we construct, there are three distinct types of edges. GATv2 enables isolated message passing on different types of edges. The input node features matrix is denoted as $\mathbf{X}=\begin{bmatrix} \mathbf{x}_\text{1} & \mathbf{x}_\text{2} & \mathbf{x}_\text{3} & \mathbf{x}_\text{4}\end{bmatrix}^\mathbf{T}$, the graph connectivity information as $\mathbf{A}$, and edge types as $e, e\in \lbrace for,against,vote\rbrace$. The neighbors set of $v_i$ under $e$ is denoted as $\mathcal{N}_{i,e}$, the weight matrix as $\mathbf{W}$, and the attention parameter between $v_i$ and $v_j$ under $e$ is $\alpha_{i,j,e}$. The process of GATv2 is formally described as follows:
\begin{equation}
	\begin{split}
		& \mathbf{h}_{i,e}^{(l)} = \sigma \left(\textstyle \sum_{j \in \mathcal{N}_{i,e}^{(l)}}^{} \alpha_{i,j,e}^{(l)} \cdot \mathbf{W}_{e}^{(l)} \mathbf{x}_j^{(l)}\right),\\
		& \alpha_{i,j,e}^{(l)} = \\
		&\frac{\exp(a_{e}^{(l)T} \text{LReLU} (\mathbf{W}_{e}^{(l)} \cdot [\mathbf{h}_{i,e}^{(l)} \lVert \mathbf{h}_{j,e}^{(l)}]))}{\sum_{k \in \mathcal{N}_{i,e}^{(l)}} \exp(a_{e}^{(l)T} \text{LReLU} (\mathbf{W}_{e}^{(l)} \cdot [\mathbf{h}_{i,e}^{(l)} \lVert \mathbf{h}_{k,e}^{(l)}]))},\\
		&\mathbf{x}_{i}^{(l+1)}=aggr({\mathbf{h}_{i,e}^{(l)}}),
	\end{split}
\end{equation}
where $l$ denotes the layer index, and $a$ is a learnable attention weight vector. $\text{LReLU}(\cdot)$ is a LeakyReLU function. The output of the second GATv2 layer is $\mathbf{X}^{(2)}$, which we first concatenate into the vector $\mathbf{X}_1= [\mathbf{x}_{1}^{(2)},\mathbf{x}_{2}^{(2)},\mathbf{x}_{3}^{(2)},\mathbf{x}_{4}^{(2)}]$. This is followed by ReLU activation function and a dropout layer \cite{velickovic2018graph}. The final output $\mathbf{\hat{y}}$ is obtained through a linear layer with a dimensionality of 4 and a softmax layer, i.e., $\mathbf{\hat{y}} \in \mathbb{R}^{1 \times 4}$. The ground truth is $\mathbf{y} \in \mathbb{R}^{1 \times 4}$. When training the expert model, a cross-entropy loss function is imposed, i.e., 
\begin{equation}
	\mathcal{L}=\mbox{Cross\_Entropy\_Loss}(\mathbf{y},\mathbf{\hat{y}}).
\end{equation}

\subsubsection{Inference of LLM-driven Judge Agent}

\begin{figure*}[t]
	\centering
	\includegraphics[width=0.8\textwidth]{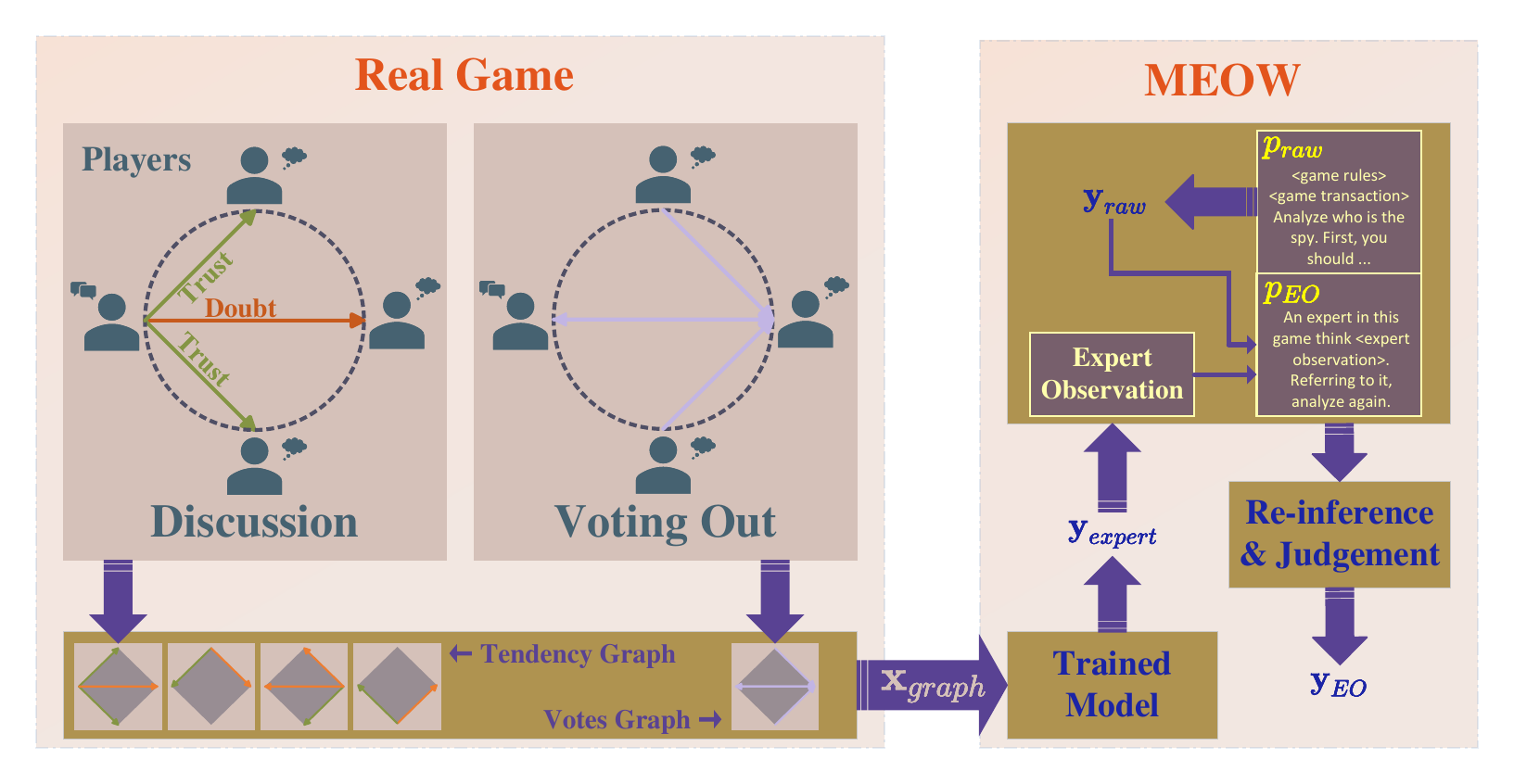}
	\caption{The framework of MEOW in the four-player version of ``Find The Spy''.}
	\label{fig:EO}
\end{figure*}

With the expert model $M_e$ trained on simulated data, an LLM-driven judge agent gets access to expert observations in real games, even when their initial states differ from those in our simulated games. As illustrated in Figure \ref{fig:EO}, the data of interaction behaviors and players' states recorded in text form are added to the initial prompt, represented as $p_{raw}$, which allows the judge agent to infer which player is the ``spy'' for the first time. We denote the inference result as $\mathbf{y}_{raw}$. Concurrently, the real data are transformed into graph-structured data $\mathbf{x}_{graph}$ using the same method as in simulated data and then inputted into the expert model. Following the judge agent's initial inference, the output $\mathbf{y}_{expert}$ of the expert model is transformed into prompts $p_{EO}$, which instruct the judge agent to make another inference. We anticipate that the weighted average F1-score between the refined inference result $\mathbf{y}_{EO}$ and the ground truth $\mathbf{y}$ is the lowest, i.e., 
\begin{equation*}
	\begin{split}
		&f(\mathbf{y}_{true}, \mathbf{y}_{pred})=\mbox{WA\_F1}(\mathbf{y}_{true}, \mathbf{y}_{pred}),\\
		&f(\mathbf{y}, \mathbf{y}_{EO}), f(\mathbf{y}, \mathbf{y}_{expert}) \ge f(\mathbf{y}, \mathbf{y}_{raw}).
	\end{split}
\end{equation*}

\section{Experiment}

\subsection{The Choice of LLM}

During the demo design stage of our experiment, internLM-20B \cite{2023internlm} was as our language model. Despite its superior performance over ChatGPT on major evaluation datasets such as RACE \cite{2017-race}, it still struggled to reason effectively with its common sense obtained through pre-training. The agents driven by it failed to comprehend the rules provided by our prompts and tend to repeat the same answer to the same questions about different players, mirroring their initial responses. Therefore, to balance cost and effectiveness, we opt for the latest ERNIE-Bot-8k \cite{wang2021ernie} as the LLM model used in the experiments. Since ERNIE-Bot-8k exhibits superior reasoning ability in Chinese compared to English, we design the prompts in Chinese. The examples in this paper are summarized and translated from Chinese.

\subsection{Agent Response}

Reviewing the simulation log, we confirm that agents driven by ERNIE-Bot-8k have the ability to understand the game rules and make rational analyses of other players' utterances and behaviors. For example, receiving the word ``sunrise'', the player agent Alice noticed an anomaly in Daniel's description.
\begin{verbatim}[fontsize=\small, commandchars=\\\{\}]
\textbf{User:} In this round, the fourth player 
Daniel's decription of his word is "light 
fades". <prompts instructing Alice to make
analysis based on common sense>
\textbf{Assistant:} <Alice's analysis>
...
(discussion process starts)
...
\textbf{User:} Reviewing utterances in this round,
analyze other player's motivation.
\textbf{Assistant:} <analysis of Bob and Carol>
Daniel describes "light fades", which I 
think is more likely the "Spy", since the 
sky gradually lights up at sunrise, rather 
than darkens.
\end{verbatim}
Moreover, still in this game, Bob received the same word as Daniel which means Alice was the ``Spy''. After Alice doubted Daniel in statement, Bob raised the opposite doubt.
\begin{verbatim}[fontsize=\small, commandchars=\\\{\}]
\textbf{Assistant}: <thinking Daniel received the 
same word as his>
...
(discussion process starts)
...
\textbf{User}: <(broadcast) Alice's statement 
(doubting Daniel)>
\textbf{Assistant}: <analysis of other players 
(trusting Daniel)> For Alice, I am against 
her. Her doubts about Daniel may have been 
strategic to increase her credibility. I 
think that during the game, players can 
be strategic and deceptive in their 
actions and speech.
\end{verbatim}
\par These two agent response samples illustrate the feasibility of simulating communication games with generative agents, and the agents are able to reason with knowledge like human players during the simulated game.

\subsection{Dataset}
We utilize the aforementioned dataset construction method in Section \ref{sec:DC}. Under diverse settings of words players receive, the single-round dataset $D_1$ consists of 261 valid heterogeneous graphs, totaling 1044 nodes and 3551 edges. The two-round dataset $D_2$ contains 181 valid heterogeneous graphs, amounting to 724 nodes and 3966 edges. To ensure the stability of our experiments, we implement ten different random data split configurations, yielding ten distinct training and test sets that respectively represent simulated games and real games. A summary of the statistical features of these datasets is presented in Table \ref{tab:dataset}.

\begin{table}
	\centering
	\begin{tabular}{l ccc}
		\toprule
		dataset & graphs & nodes & edges \\
		\midrule
		$D_{1}$ & 261 & 1044 & 3551 \\
		$D_{2}$ & 181 & 724 & 3966 \\
		\bottomrule
	\end{tabular}
	\caption{Summary of the datasets used in our experiments.}
	\label{tab:dataset}
	\vspace{-10pt}
\end{table}

\subsection{Model Training}

The architecture of the two models is detailed in Section \ref{sec:sec_expert_model}. Each dataset is systematically partitioned into ten distinct training and test sets, each of which is used to train an expert model. To ensure robustness and generalize the model's performance across diverse data scenarios, a four-fold cross-validation strategy is adopted, and training stops if the model's performance on the validation set consistently deteriorates over a continuous span of $K$ epochs, compared to the best validation loss identified previously. 

Denoting $H_i$, $O_i$, $i\in \lbrace 1,2 \rbrace$ as the numbers of attention heads, output channels of the $i$th GATv2 layer, $\alpha$ as learning rate, $\lambda$ as weight decay rate, $p$ as dropout rate, the hyperparameters $(H_1,O_1,H_2,O_2,\alpha,\lambda,p,K)$ of the first and second expert model are respectively $(6,32,6,16,sum,0.0001,0.0005,0.5,30)$ and $(6,32,6,18,sum,0.0001,0.0005,0.5,50)$.

\subsection{Ablation Study}

To validate the effectiveness of MEOW, we perform ablation studies on three distinct inference approaches.
\begin{itemize}
	\setlength{\itemsep}{0pt}
	\item \textbf{ERNIE-Bot with Chain-of-thoughts (EB w/ CoT):} Provide game rules, hints of words, and game broadcast messages including descriptions, utterances, and votes. Then, the judge agent is prompted to infer the agents' actions step by step, following a predefined sequence of steps that includes a strategy review, utterances analysis, and motivation inference, culminating in a final identity judgment.
	\item \textbf{Expert:} The output of the expert model, trained on simulated data, i.e., the expert observation, is considered as the final identity judgment.
	\item \textbf{ERNIE-Bot with Chain-of-thoughts and Expert Observation (EB w/ CoT \& EO):} Following the same process as EB w/ CoT, the judge agent is prompted to re-infer the game situation and determine whether to adjust the final judgment based on the expert observation.
\end{itemize}

\begin{table}
	\centering
	\begin{tabular}{l ccc}
		\toprule 
		Method & Round & Acc. & WA-F1 \\ 
		\midrule 
		EB w/ CoT & 1 & 30.66 & 29.79 \\
		Expert & 1 & 31.60 & 31.36 \\
		EB w/ CoT \& EO & 1 & 30.43 & 30.59 \\
		\midrule
		EB w/ CoT & 2 & 28.72 & 28.96 \\
		Expert & 2 & 38.85 & 38.01 \\
		EB w/ CoT \& EO & 2 & 36.82 & 36.50 \\
		\bottomrule 
	\end{tabular}
	\caption{Ablation studies conducted on MEOW. `Acc.' denotes the accuracy score, and `WA-F1' represents the weighted average F1-score. In the results generated by EB, responses such as ``I'm not certain'' are considered random predictions, with their randomness determined based on multi-turn dialogue logs.}
	\label{tab:AS}
	\vspace{-10pt}
\end{table}

For the purpose of avoiding the advantage of ``Expert'' on training sets, ablation studies are exclusively conducted on the test sets. We perform ten independent experiments with distinct randomized training and test sets to guarantee the reliability of our results. The average statistic of the eight sets, representing the median performance among the ten sets, is presented in Table \ref{tab:AS}.

In the first round, ``EB w/ CoT'' attains an accuracy score of 30.66 and a WA-F1 score of 29.79. ``Expert'' outperforms by approximately 1 percent, achieving an accuracy score of 31.60 and a WA-F1 score of 31.36. The combined method ``EB w/ CoT \& EO'' registers an accuracy score of 30.43 and a WA-F1 score of 30.59, approximating the average of ``EB w/ CoT'' and ``Expert''. Similarly, in the second round, ``EB w/ CoT'' keeps a better performance than ``Expert'', and ``EB w/ CoT \& EO'' remains intermediate to the two methods. In both rounds, the Expert model exhibits superior performance compared to the CoT methods, and expert observation effectively aids in correcting some errors in the initial inference, as reflected in the statistics.

\begin{table}
	\centering
	\begin{tabular}{l cc}
		\toprule 
		Method & Round 1 & Round 2 \\ 
		\midrule 
		EB w/ CoT & 0.0419 & 0.0635 \\
		Expert & 0.0298 & 0.0232 \\
		EB w/ CoT \& EO & 0.0146 & 0.0288 \\
		\bottomrule 
	\end{tabular}
	\caption{Standard deviation of Weighted-average F1 score in round 1 and round 2.}
	\label{tab:STDEV}
	\vspace{-10pt}
\end{table}

Besides, as is shown in Table \ref{tab:STDEV}, after refinement of expert observations, the standard deviation of WA-F1 score reduces significantly. This indicates that MEOW can enhance the stability of LLM agents' reasoning while improving its performance.

\section{Discussion}

Through our experiments, training an expert model based on simulated data is both feasible and beneficial for addressing problems in complex human systems. In the absence of any conversation information in a new game, expert models outperform manual CoT methods. On one hand, simulation expands the available data of states in a system, enabling models trained on these data to ``experience'' more, thereby refining LLM inference even without any historical data. On the other hand, in complex human systems, current LLMs with CoT prompts are not capable of fully comprehending human behaviors and other system forms. As a result, in our two-round experiments, ``LLM w/ CoT'' fails to recognize that the player voted out in the first round cannot be the ``spy'', and its performance
\begin{verbatim}[fontsize=\small]
The results of the first round of voting 
are: Alice votes for Carol, Bob votes for 
Alice, Carol votes for Bob, and Daniel 
votes for Bob. Thus, Bob is eliminated.
...
Next up is player Bob, who I consider to 
be the "spy".
\end{verbatim}
falls below 33 percent, which is the expectation of random guessing.

When adjusting the parameters of the judge agent, we discover that at the setting of $(top\_p, temperature)=(0.85, 0.8)$, the judge agent achieves an optimal balance between sticking to itself and illogical inferences. In this setting, expert observations do correct some inferences.
\begin{verbatim}[fontsize=\small]
According to the judgment of game experts, 
Daniel is likely to be the "spy". I think 
there is some truth to it, because Daniel 
did not give any clear words related to 
bees in the description and discussion 
session, and his descriptor "unique" was 
relatively vague, which is consistent with 
the characteristics of the "spy" trying 
to hide his identity.
\end{verbatim}

Therefore, MEOW is never the substitute for FT, CoT, ICL, and RAG, but rather to collaborate with them to enhance the performance of inference in complex human systems. By leveraging these rapidly evolving reasoning methods, the simulation process in MEOW becomes more authentic, thereby increasing the value of expert observation.

\section{Conclusion}

In this paper, we propose a novel methodology aiming at enhancing the reasoning ability of LLMs in addressing problems concerning complex human systems through simulation. It brings unprecedented human behavior simulation and has the potential for addressing data collection challenges out of privacy, security, etc. in special scenarios. Leveraging the simulated data, we present MEOW, a framework that employs ML models to learn from these data and generate expert observations. By referring to these expert observations, LLMs can refine their inferences, typically yielding results better than those obtained using the CoT-only method. We believe that beyond the realm of communication games, in real-world complex human systems such as sociology and economics, both the generative-agents-based simulation technique and MEOW hold promise as effective methodologies. We are looking forward to the development of more powerful LLMs that can further enhance the effectiveness of our proposed framework.

\section*{Limitations}

Hallucination is the primary limitation when scaling generative-agents-simulation and subsequent MEOW. Given that LLMs generate text in a ``token by token'' manner, setting a relatively low temperature to achieve the desired diversity in simulation can lead to errors in some key tokens among all dialogues. For instance, during the discussion process, double quotation marks are followed by identities or word descriptions. If we set the temperature to $0.8$, we observe the following response in the simulation, 
\begin{verbatim}[fontsize=\small]
Player Bob's description "Spy" ...
\end{verbatim}
which results in illogical reasoning. The security of LLMs presents another limitation if we aim to utilize generative agents and MEOW in human-related security scenarios. To prevent negative generation of LLMs, token chains that violate laws and social norms are penalized during pre-training. However, this restricts the ability of generative agents to simulate offensive attackers. Cost is the final limitation, and its impact becomes more significant when scaling. Implementing generative-agent-based simulation and MEOW requires large volumes of multi-turn dialogues, with most previous turns needing to be retained as context for the next generation. This implies that after completing one generation, there are fees for its tokens in every subsequent turn. In our experiments, using GPT-4 would cost us 5400 dollars, and this is only for a four-player version. As the number of agents increases, there will be more interactions, leading to nonlinear growth in cost.

\bibliography{anthology,custom}
\bibliographystyle{acl_natbib}

\newpage
\appendix

\section{Game Processes}\label{sec:GP}

\begin{enumerate}
	\item At the beginning of a game, the referee distributes words to all players, and then several processes loop until one side wins. A loop of 2, 3, 4 is called one round, and a game may consist of one or more rounds;
	\item Description process: Players take turns to deliver statements, using a word (cannot be their own word or include all their fellow players who used their words before) to describe their received word;
	\item Discussion process: After all players give descriptions of received words, they take turns to express opinions on other players' identities.
	\item Voting process: Each player chooses a possible ``spies'' player to vote for him/her. The Player getting the most amounts of votes is eliminated from the game (If several players get the same number of votes, the player whose first vote is got earlier will be eliminated.). The referee will announce which player is eliminated by voting and reveal his/her identity. The eliminated player will be excluded from the rest of the game.
\end{enumerate}
Other rules: a) In each player's utterances at any time it is not allowed to use the word he/she receives. b) In each player's description of words in each round they must not use words used by other players before.

\section{Six Phases} \label{sec:SP}

\begin{enumerate}
	\setlength{\parskip}{0pt}
	\item \textbf{Game initialization}
	\begin{enumerate}
		\item \textbf{Game initialization phase:} Every player agent is prompted the game rules, game skills, the players in the game, its received word, and a request to confirm above information.
	\end{enumerate}
	\item \textbf{Description process}
	\begin{enumerate}
		\item \textbf{Identity inference phase:} Before describing the word, the player agent is prompted to infer its identity in the game based on other players' utterances and behaviors so as to guess the word which players of the opposite identity received.
		\item \textbf{Word description phase:} The player agent is first prompted to recall the word they received and the game rules. Consequently, it establishes its description strategy (be honest or tell a lie). Finally, we prompt the player agent to give its description in textual JSON format so that the game engine can complete the transformation efficiently, and this description is the only information open to other players in this process.
		\item \textbf{Description analysis phase:} After a player agent makes its word description, the rest player agents will be prompted to analyze it. They are requested to judge whether it is proper to describe their received words and further consider its identity.
	\end{enumerate}
	\item \textbf{Discussion process}
	\begin{enumerate}
		\item \textbf{Statement and discussion phase:} Firstly, a player agent is prompted to recall its identity and other players' descriptions and statements in this round. Secondly, it needs to choose the players it will be for and against, which will have to be embodied in his following statement. Thirdly, we prompt it to confirm its tendency in textual JSON format. Finally, it makes its statement corresponding to its tendency in this round, and this statement is the only information open to other players in this process. 
	\end{enumerate}
	\item \textbf{Voting process}
	\begin{enumerate}
		\item \textbf{Vote phase:} Player agents are prompted the vote rules and have an opportunity to rethink the situation of the current game. Finally, they are requested to make their voting in JSON format, and this is the only information open to other players in this process.
	\end{enumerate}
\end{enumerate}

\end{document}